\documentclass{article}

\usepackage[preprint,nonatbib]{neurips_2020}

\usepackage[utf8]{inputenc} 
\usepackage[T1]{fontenc}    
\usepackage{hyperref}       
\usepackage{url}            
\usepackage{booktabs}       
\usepackage{amsfonts}       
\usepackage{nicefrac}       
\usepackage{microtype}      
\usepackage{graphicx}
\usepackage{makecell}
\usepackage{tcolorbox}
\newcommand{\highlight}[1]{\colorbox{blue!10}{#1}}

\title{{\normalfont\textit{Jigsaw}-}VAE: Towards Balancing Features \\ in Variational Autoencoders}

\author{
  Saeid Asgari Taghanaki \\
  Autodesk AI Research\\
  Toronto, Canada 
  \And
  Mohammad Havaei\thanks{Equal contribution} \\
  Imagia \\
  Montreal, Canada 
  \And
  Alex Lamb\footnotemark[1]\\
  MILA \\
  Montreal, Canada 
  \And
  Aditya Sanghi \\
  Autodesk AI Research \\
  Toronto, Canada 
  \And
  Ara Danielyan \\
  Autodesk AI Research \\
  Toronto, Canada 
  \And
  Tonya Custis \\
  Autodesk AI Research \\
  San Francisco, USA 
}

\begin{document}

\maketitle

\begin{abstract}
The latent variables learned by VAEs have seen considerable interest as an unsupervised way of extracting features, which can then be used for downstream tasks.  There is a growing interest in the question of whether features learned on one environment will generalize across different environments. We demonstrate here that VAE latent variables often focus on some factors of variation at the expense of others - in this case we refer to the features as ``imbalanced''. Feature imbalance leads to poor generalization when the latent variables are used in an environment where the presence of features changes. Similarly, latent variables trained with imbalanced features induce the VAE to generate less diverse (i.e. biased towards dominant features) samples. To address this, we propose a regularization scheme for VAEs, which we show substantially addresses the feature imbalance problem. We also introduce a simple metric to measure the balance of features in generated images. 
\end{abstract}

\begin{figure}[htb!]
    \centering
    \includegraphics[width=\textwidth]{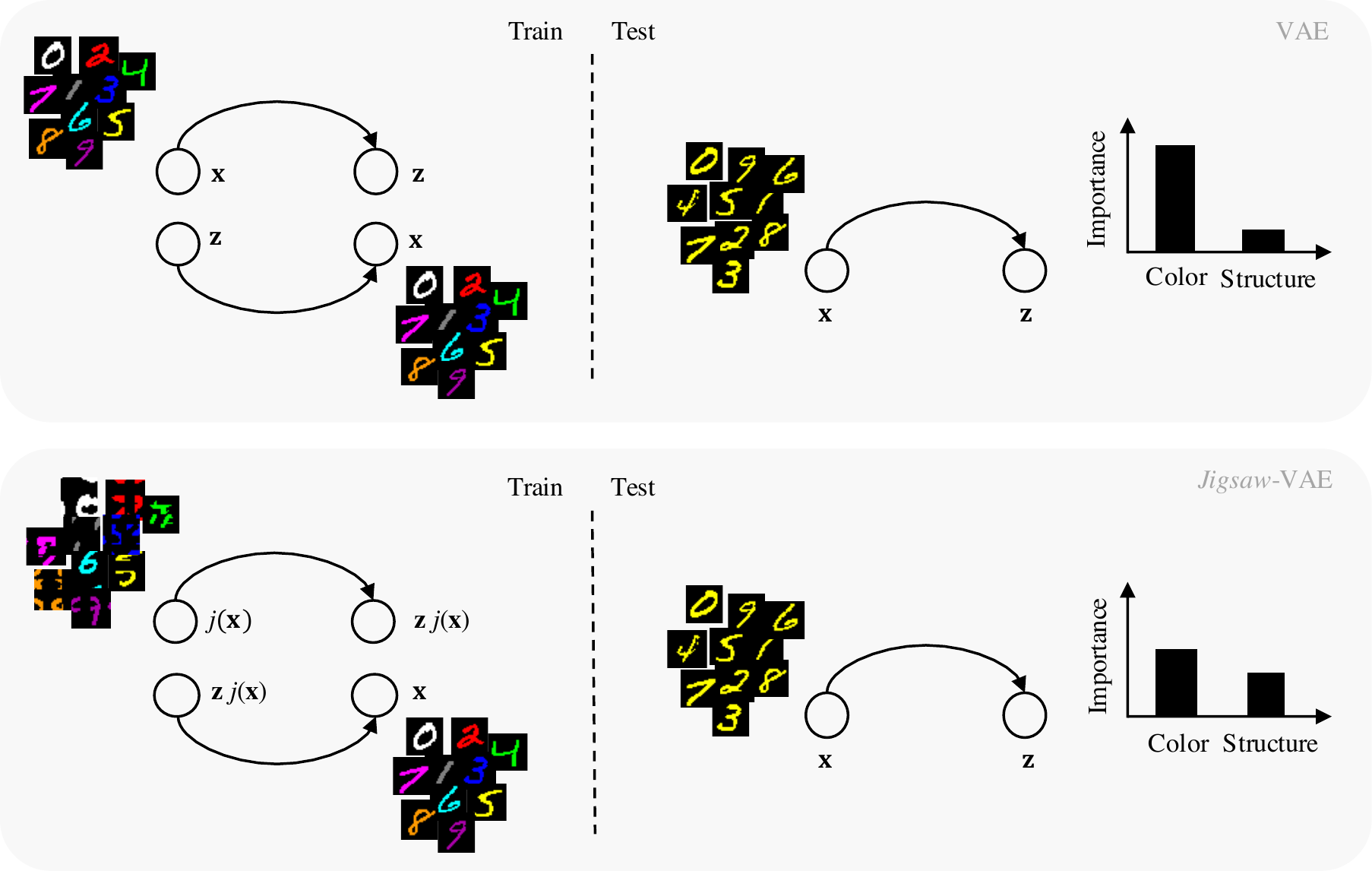}
    \caption{Colored MNIST example of balancing features : comparison between the VAE and our proposed \textit{Jigsaw}-VAE. Note that the permutation function $J$ is applied on each individual sample separately.}
    \label{fig:introfig}
\end{figure}

\section{Introduction}
Variational autoencoders (VAEs) have certain appealing properties compared to generative adversarial networks (GANs), such as stable training~\cite{tolstikhin2017wasserstein}, interpretable inference~\cite{brock2016neural}, and calculating data likelihood~\cite{kingma2013auto}. As such, they remain worthy for examination and improvement. Additionally, recent VAEs~\cite{razavi2019generating} have shown a great potential for generating competitive high quality images compared to the state-of-the-art GANs. 

However, there are a few identified challenges with regard to VAEs: balancing the two terms in the VAE loss function (i.e. the log likelihood and the Kullback–Leibler (KL) divergence terms) is not trivial. Sacrificing the former causes sub-optimal reconstruction performance while neglecting the latter results in poor sampling (several latent variables might be ignored/over-pruned~\cite{yeung2017tackling}). To tackle this, several works have explored weighting the terms in a more systematic way than the original VAE e.g. by annealing the contribution weight~\cite{bowman2015generating,sonderby2016ladder} or by learning it~\cite{asperti2020balancing,dai2019diagnosing}. Additionally, even after convergence, a matched latent prior with a learned aggregated posterior distribution is not guaranteed, which can be due to the choice of an overly simplistic prior distribution~\cite{kingma2016improved,dai2019diagnosing}. A simple prior and/or improperly weighted evidence lower bound (ELBO) terms can lead to less diverse generated samples~\cite{tomczak2017vae,higgins2017beta} i.e. neglecting the minor (sub-) clusters of input samples/features.

In this paper, we study VAEs from a slightly different perspective. We examine whether optimized VAEs via the ELBO tend to ignore sporadic features of input data since the models are able to optimize the ELBO faster by comfortably favoring the optimization towards more frequent, dominant features. Handling \textit{feature imbalance} (a.k.a representation bias~\cite{li2018resound,suresh2019framework} in general) in variational models, particularly VAEs, seems to be more complex than supervised and non-variational  models~\cite{zhao2017men,hendricks2018women,leino2018feature,kim2019learning, hardt2016equality,lorenz2019unsupervised,clark2019don,li2019repair} where they have access to the prior information/distribution of features or they specify a target feature (e.g. color) to balance. However, it is not trivial to find problematic features in a large dataset to target while performing a generative task or representation learning. Similarly, the classic problem of class imbalance in supervised classification models can be handled with several different approaches such as by simply adding penalty terms/weights to the model's loss function to enforce the model for a fair, balanced prediction. However, when it comes to generative models and representing learning, there can be many large and small clusters of samples in training data (usually without fine-grained cluster labels). Here, the question to be answered is: \textit{what does imbalance mean in generative models and representation learning and how it can be handled?} In other words, for the same example above, how do we enforce a generative model to not ignore a non-dominant cluster while generating random samples? Note that even enforcing the model to handle the high level imbalance of ``female'' vs ``male'', the same issue can be exacerbated by extending to sub-clusters of each of these categories e.g. females with and without eyeglasses, blond or black hair.   

Therefore, besides the main challenges of VAEs discussed above, feature imbalance might also be a reason for capturing poor latent information which results in generating less diverse samples as well as under-performing in downstream tasks which are conditioned on latent features. For instance, if we assume a data distribution is defined by two main features of ``color'' and ``structure'', we would ideally want a model to learn ``equal'' contribution of the two features. However, this does not always happen in practice, which might result in a drastic failure in downstream tasks. As an example, a simple case scenario has been visualized in Figure~\ref{fig:introfig} (top) where modification of dominant feature (color) results in a feature-imbalanced latent representation. As a remedy, we apply feature permutation (Figure~\ref{fig:introfig} (bottom)) to the VAE to reduce reliance on a single or a few dominant feature(s). 

Feature permutation has previously been used as a metric to calculate feature importance in Random Forests~\cite{breiman2001random}. A feature is ``important'' if permuting its values decreases the model performance, indicating that the model had relied on that feature for its prediction. Based on this idea, Fisher et al.~\cite{fisher2019all} proposed a model-agnostic version of the feature importance. Recently, permutation based techniques, particularly \textit{Jigsaw} puzzle approaches, have been successfully applied to (semi-) supervised deep models~\cite{noroozi2016unsupervised,santa2017deeppermnet,paumard2018jigsaw} with the goal of capturing rich contextual information. However, their aptitude for addressing the feature imbalance in the latent space of VAEs and how these latent spaces subsequently affect the downstream tasks has not yet been studied. In this paper, we focus on \textit{whether permutation-based regularization is capable of reducing the adverse effect of feature imbalance in VAEs}. 

To solve a jigsaw puzzle, a VAE has to focus on how different input variables can be integrated (sorted) to build a plausible structure. In other words, instead of reconstructing the ``big picture'' by focusing on dominant features which might be a simpler task to optimize the ELBO, the model focuses more on smaller areas that include local structural features. This prevents the model to choose the easy optimization path of learning \textit{only} dominant variables. 

In this paper, we make the following contributions:

\begin{itemize}
    \item We inspect the feature imbalance issue in the context of VAE. We study whether the VAE learns balanced features given feature-imbalanced input data, which is more often the case in practice. Training a VAE with feature-balanced data is the ideal way of achieving this goal. However, collecting feature-balanced data requires great effort.  \item We propose a simple metric called \textit{Feature Presence Metric} to measure the balance of features in generated images.
    \item We propose a simple yet effective feature  permutation-based technique to systematically enforce the VAE to learn balanced features. 
    \item On the line of research of improving the VAE using more flexible priors, we introduce a prior which is a mixture of a uniform permutation distribution and a Gaussian distribution.
\end{itemize}

\section{Method}

In vanilla VAE, the prior distribution $p(\textbf{z})$ is defined on the latent representation $\mathbf{z} \in \mathbb{R}^{D}$ and is usually set to an isotropic Gaussian distribution $\mathcal{N}\left(0, I \right)$ where $D$ is the latent dimension. The posterior distribution is defined as $p_{\theta}(\mathbf{z} | \mathbf{x}) \propto p_{\theta}(\mathbf{x} | \mathbf{z}) p(\mathbf{z})$. Then a parameterized distribution over $\mathbf{x}$ is defined as $p_{\theta}(\mathbf{x} | \mathbf{z})$ and is modeled as a generative network in the context of neural network such that $\theta$ becomes the weights of the network. Similar to the generative network, a neural network is used to approximate distribution $q$ conditioned on observation $\mathbf{x}$, called inference network $q_{\phi}(\mathbf{z} | \mathbf{x})$ with variational parameter $\phi$ which is also weights of the neural network. Using re-parameterization trick~\cite{kingma2013auto} back-propagation is applied on the parameter $\phi$ considering $\textbf{z}$ as a function of noise and typically mean and variance of the Gaussian learned by a decoder network. Therefore, the objective of VAE is to maximize the following variational lower bound with respect to the parameters $\theta$ and $\phi$,

\begin{equation}
    \log p_{\theta}(\mathbf{x}) \geq \mathbb{E}_{q_{\phi}(\mathbf{z} | \mathbf{x})}\left[\log \frac{p_{\theta}(\mathbf{x}, \mathbf{z})}{q_{\phi}(\mathbf{z} | \mathbf{x})}\right] =\mathbb{E}_{q_{\phi}(\mathbf{z} | \mathbf{x})}\left[\log p_{\theta}(\mathbf{x} | \mathbf{z})\right]-\mathbb{K} \mathbb{L}\left(q_{\phi}(\mathbf{z} | \mathbf{x}) \| p(\mathbf{z})\right)
\end{equation}

Let $q_{\phi}(\mathbf{z} | j(\mathbf{x}))$ be a conditional normal distribution such that 

\begin{equation}
    q_{\phi}(\mathbf{z} | j(\mathbf{x})) = \mathcal{N}\left(\mathbf{z} | \mu_{\phi}(j(\mathbf{x})), \sigma_{\phi}(j(\mathbf{x}))\right).
\end{equation}

 with $\mu_{\phi}(j(\mathbf{x}))$ and $\sigma_{\phi}(j(\mathbf{x}))$ as non-linear functions and $j$ as a permutation function. If we consider $p(j(\mathbf{x})|\mathbf{x}) = \mathcal{U}_{[0, b]}(\mathbf{x})$ to be a permutation model (Jigsaw in our case) of $\mathbf{x}$ where $b$ is the number of different permutations in the uniform distribution, then
 
 \begin{equation}
      q_{\phi}^j(\mathbf{z} | \mathbf{x})=\mathbb{E}_{p(j(\mathbf{x}) | \mathbf{x})}\left[q_{\phi}(\mathbf{z} | j(\mathbf{x}))\right]=\int_{j(\mathbf{x})} q_{\phi}(\mathbf{z} | j(\mathbf{x})) p(j(\mathbf{x}) | \mathbf{x}) d j(\mathbf{x})
 \end{equation}

is a mixture of normal and uniform distributions (i.e. each time we sample $j(\mathbf{x}) \sim p(j(\mathbf{x}) | \mathbf{x})$ and feed into $q(\mathbf{z} | j(\mathbf{x}))$ we get different distributions). In practice, the inference network will learn which distribution is needed for a given $j(\mathbf{x})$.

Similar to~\cite{im2017denoising}, if we consider $q_{\phi}^j(\mathbf{z} | \mathbf{x}))$ as the approximate distribution, the variational lower bound can be written as 
\begin{equation}
    \mathcal{L}_{Jigsaw\textit{-VAE}} = \mathbb{E}_{q^j_{\phi}(\mathbf{z} | \mathbf{x})}\left[\log \frac{p_{\theta}(\mathbf{x}, \mathbf{z})}{q_{\phi}(\mathbf{z} | j(\mathbf{x}))}\right].
\end{equation}

\begin{figure}[htb!]
    \centering
    \includegraphics[width=\textwidth]{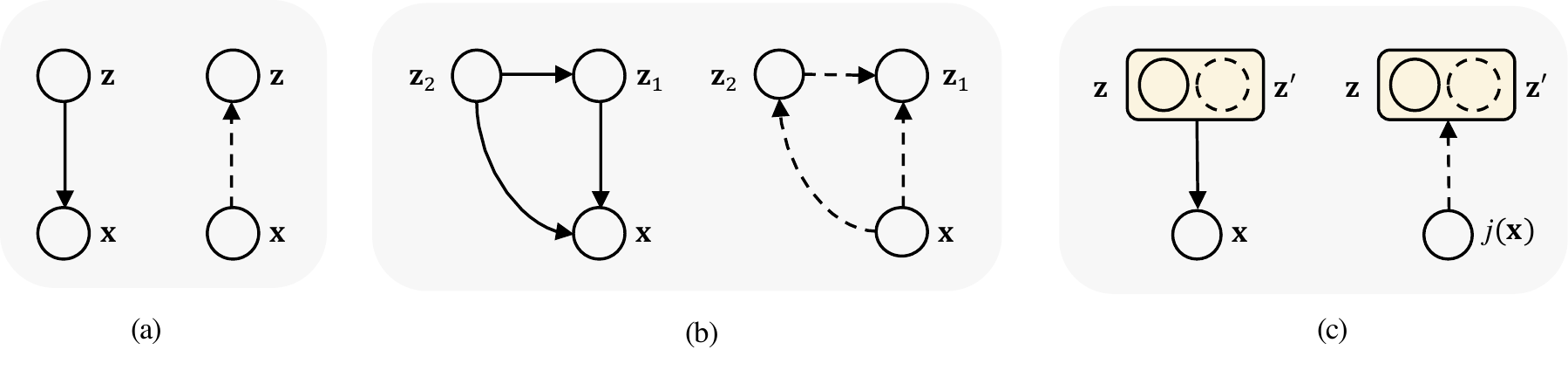}
    \caption{Stochastical dependencies. (a) VAE~\cite{kingma2013auto}, (b) VampPrior VAE~\cite{tomczak2017vae}, (c) \textit{Jigsaw}-VAE where $\mathbf{z} \sim q(\mathbf{z}|\mathbf{x})$ and $\mathbf{z^{\prime}}\sim q(\mathbf{z}|j(\mathbf{x}))$. Solid and dashed arrows show generative and inference steps, respectively.}
    \label{fig:model}
\end{figure}

Compared to $q_{\phi}(\mathbf{z} | \mathbf{x})$ in VAE, $q_{\phi}^j(\mathbf{z} | \mathbf{x})$ in \textit{Jigsaw}-VAE has the capacity to cover a broader class of distributions, thus creating more sub-spaces to decode diverse samples without intermixing sub-spaces. The comparison of the \textit{Jigsaw}-VAE with the VAE and and other models which use complex priors such as VampPrior VAE~\cite{tomczak2017vae} is depicted in Figure~\ref{fig:model}. As can be seen, the VampPrior VAE builds a hierarchical variational inference module by explicitly adding auxiliary latent vectors ($\mathbf{z}_1$, $\mathbf{z}_2$) while our \textit{Jigsaw}-VAE has a single latent vector which caries information about both $\mathbf{z}$ and $\mathbf{z^{\prime}} = \mathbf{z}(j(\mathbf{x}))$.

Compared to mixture of Gaussians prior and VampPrior, \textit{Jigsaw}-VAE does not have any extra parameters beyond that required by the VAE to be learned. In the case of mixture of Gaussians, $p\left(\mathbf{z}\right)=\frac{1}{K} \sum_{k=1}^{K} \mathcal{N}\left(\mu_{k}, \textrm{diag}\left(\sigma_{k}^{2}\right)\right)$ where $\mu_{k} \in \mathbb{R}^{M}, \sigma_{k} \in \mathbb{R}^{M}$ are learned parameters.

\begin{figure}[htb!]
    \centering
    \includegraphics[scale=0.6]{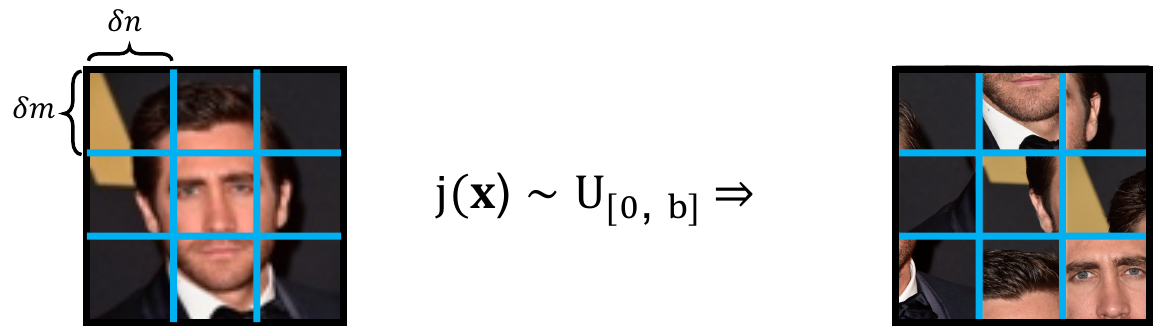}
    \caption{Sampling permutations form a uniform distribution.  $(\delta n, \delta m)$ show the window size $S$. In our experiments, we set $\delta n = \delta m= \frac{I}{4}$ where $I$ corresponds to size of a side of the square input image. Therefore the range of the permutation function $j$ will be $[0, b=16!]$ where $b$ is the total number of permutations.}
    \label{fig:permutation_function}
\end{figure}

The permutation function we use as the input stochastic layer is depicted in Figure~\ref{fig:permutation_function}. The function can be defined over spatial windows $S$ with size $(\delta n, \delta m)$, color channels, or both.

\section{Related Work}

\textbf{Smoothing the conflict between the ELBO terms.} Ideally, we want every data sample to be (a) encoded into a latent space such that it can be accurately reconstructed and (b) close enough to the origin that we are likely to hit that point in latent space by sampling $\mathbf{z} \sim \mathcal{N}(0, I)$. Note that (a) and (b) are somewhat in conflict~\cite{higgins2017beta,song2019latent}: enforcing the latent points to be far from each other makes (a) easier but harms (b). This conflict can be softened either by defining more complex priors than the Gaussian as done in VampPrior VAE~\cite{tomczak2017vae} and denoising VAE~\cite{im2017denoising}, or by modifying the VAE objective~\cite{higgins2017beta}. Our proposed \textit{Jigsaw}-VAE adds an extra stochastic sampling layer to the beginning of the encoder which makes the latent sampling space a mixture of distributions. However, in comparison with works~\cite{tomczak2017vae,im2017denoising} which use more flexible priors than the standard Gaussian, our method does not need to modify the network architecture, nor does it need extra training with sudo inputs as done by Tomczak et al.,~\cite{tomczak2017vae}. In contrast to methods that modify the ELBO by introducing a weighting coefficient for the KL term~\cite{higgins2017beta,bowman2015generating,sonderby2016ladder,asperti2020balancing,dai2019diagnosing}, our method does not require any balancing factor to determine the contribution of the likelihood and the KL term in the overall optimization.

\textbf{Stochastic layer in the bottom of encoder.} Compared to denoising VAE (\textit{d}-VAE)~\cite{im2017denoising} which adds an extra stochastic noise layer to the VAE, we use a jigsaw permutation function based layer to systematically enforce the capturing of structural features by the VAE. Our jigsaw approach encodes the prior assumptions into the model such that the structure of the world (relative position of parts) is more crucial to get right than the pixel-level appearance which it is done by \textit{d}-VAE.

\textbf{Enforced information placement in latent vector.} The work done by Chen et al.,~\cite{chen2016variational} is related to our method in the sense that they have also induced the latent space to carry some desired information. To be specific, however, we make the decoding distribution of our method to be incapable of modeling information that the Jigsaw-VAE's encoder captures (i.e. thanks to the permutation function it codes the local spatial relationships). The vector quantized VAE~\cite{razavi2019generating} also filters the information in the latent space such that only the most discriminative features are passed through the decoder. Louizos et al.,~\cite{louizos2015variational} have proposed a penalty term based on maximum mean discrepancy to obtain purged latent information with removed unwanted sources of variation. However, we prefer not to remove infrequent useful features. Instead, our method balances and recovers infrequent features in the latent space. Our work also differs from disentanglement approaches~\cite{mathieu2018disentangling,kim2018disentangling}. Although they learn independent factors of variations, they do not guarantee that the less frequent factors will not be ignored or frequent features will be dominated. 

In contrast to all of the above, to the best of our knowledge, our work is the first to study the feature imbalance concept in VAEs and its effect on downstream tasks. It is also the first work to explore the effect of a \textit{permutation}-based stochastic layer, placed at the bottom of the VAE's encoder.

\section{Experiments}

In the experiments we aim at verifying empirically
whether the \textit{Jigsaw} approach (a) helps the VAE to learn a representation that can preserve rare features after convergence using ELBO in generative tasks, and (b) whether it helps learning balanced features for downstream tasks such as clustering.

Therefore we group the experiments into two main categories. To answer (a), in subsection~\ref{subsec:diversity}, using our proposed metric (i.e. feature presence metric: subsection~\ref{subsec:fpm}) we inspect the presence of targeted features in a generative task; to answer (b), in subsection~\ref{subsec:clustering}, we measure the discriminate performance of the latent information in a  feature-biased clustering scenario. 

\subsection{Feature Presence Metric} \label{subsec:fpm}
Various metrics have been proposed to evaluate the performance of VAEs. Among them, reconstruction error, negative log likelihood, Inception score (IS)~\cite{salimans2016improved} and Frechet Inception distance (FID)~\cite{heusel2017gans} (or a collection of these) are generally used. IS and FID are both based on a model trained on Imagenet~\cite{deng2009imagenet} and as reported before, they do not properly capture fidelity or diversity~\cite{barratt2018note} (e.g. unrealistic images can obtain high IS and FID scores). We argue that in addition to the above metrics (regardless of their limitations), measuring whether features/variables of the train set are preserved after a VAE is converged is an important factor towards both robust representation learning and diversity in generated samples. To this end, we propose the feature presence metric (FPM) which can be calculated over randomly generated images of a generative model. FPM measures how well a particular feature/variable is perpetuated in a set of generated images. 

For a target feature $f$, FPM is calculated as: 
 
\begin{equation}
    FPM = \left |(\frac {N_{gf}} {N_g}) - (\frac {N_{tf}} {N_t}) \right | \times 100
\end{equation}

where $N_{gf}$ and $N_{tf}$ are the number of randomly generated and train (real) images which contain the target feature, respectively. $N_g$ and $N_t$ are the total number of the generated and train images, respectively. To calculate $N_{gf}$ we train a classifier which detects whether the targeted feature is present in a sample. 

\subsection{Feature Inspection}\label{subsec:diversity}

In this section, we use the CelebFaces Attributes (CelebA) dataset~\cite{liu2015deep} to train the models and analyze the diversity of the images randomly generated by each method. We compare our \textit{Jigsaw}-VAE to four other VAE methods: the vanilla VAE~\cite{kingma2013auto}, \textit{d}-VAE~\cite{im2017denoising} $\beta$-VAE~\cite{higgins2017beta}, and the VampPrior VAE~\cite{tomczak2017vae}. Inspired from the Mixup approach ~\cite{zhang2017mixup} which was applied as a data augmentation in supervised tasks, we train the vanilla VAE with mixed inputs (\textit{Mixup}-VAE) as another baseline. Particularly, we add stochasticity to the input of the VAE by mixing training samples using the linear interpolation mentioned in~\cite{zhang2017mixup}. We then set the ground truth for the VAE to be one of the mixed samples which has the max contribution in the input mixed-up sample.

To evaluate the methods' generative performance in terms of FPM, we train MobileNet-V2~\cite{sandler2018mobilenetv2} to predict the attributes of each randomly generated image. We choose a diverse set of features including both under- and over-represented ones i.e. features that are present in only $\sim 2\%$ to $\sim 58\%$ of the training data (Table~\ref{tab:celeb}). We generate 5000 random samples per method and report the frequency of the targeted attributes present in the generated images. 

\begin{figure}[htb!]
    \centering
    \includegraphics[width=\textwidth]{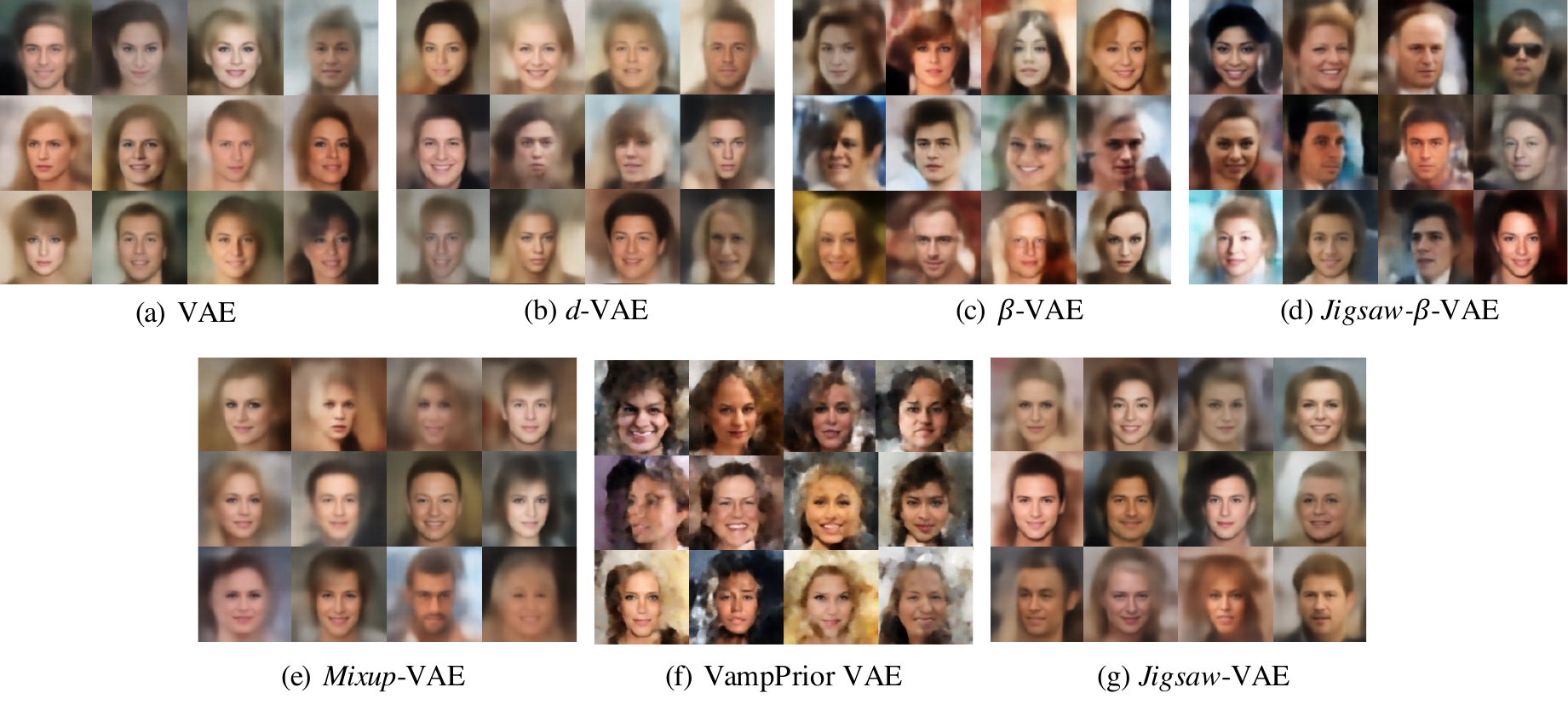}
    \caption{Randomly generated samples of size $128 \times 128$ with different methods; zoom for details.}
    \label{fig:samples}
\end{figure}

In Table~\ref{tab:celeb}, we report the performance of each method on imbalanced features i.e. we measure whether each method is capable of preserving the balance of features. Ideally, FPM values for each feature should be zero. As can be seen the \textit{Jigsaw}-VAE achieves favorable MSE and FPM values. $\beta$-VAE and VampPrior VAE obtain better values for some of the under-represented attributes as well as reconstruction error. However, looking at Figure~\ref{fig:samples}, the random images generated via $\beta$-VAE and VampPrior VAE seem less realistic than our proposed \textit{Jigsaw}-VAE. After applying our Jigsaw approach to $\beta$-VAE (i.e. \textit{Jigsaw}-$\beta$-VAE), the generated images look more realistic compared to $\beta$-VAE and achieve better FPM values. Although the VampPrior VAE leverages more complex latent calculations, it does not scale to high dimensional images; as shown in Figure~\ref{fig:samples}-(f). Considering FPM values and whether the randomly generated images look realistic, jigsaw based approaches outperform others. This experiment showed that the $\beta$-VAE is able to preserve under-represented features to some extent, but it generates unrealistic images. However, applying the jigsaw approach to both VAE and $\beta$-VAE leads to preserved underrepresented features as well as generating more realistic images.

\begin{table}[htb!]
\centering
\caption{Reconstruction error (i.e. MSE) and FPM values (lower is better) for features from celebA dataset. Numbers in parenthesis below each feature name are percentage of the data that has the feature. In each column, two best values are highlighted.} 
\begin{tabular}{l|c|cccccc|c}
\hline
Method                                      & MSE     & \thead{Female\\(58.06)}  & \thead{Eyeglasses\\(6.46)}     & \thead{Bald\\(2.28)}   & \thead{Beard\\(16.58)}   & \thead{Smiling\\(47.97)}  & \thead{Gray\\(4.24)} & \thead{AVG} \\ \hline
VAE~\cite{kingma2013auto}                   & 0.0272 & 6.30    & 6.24       & 2.04  & 14.22  & 14.31    &2.76  &9.762  \\
\textit{d}-VAE~\cite{im2017denoising}       & 0.0271 & 17.52    & 6.36       & 2.10  & 13.34  & 17.00    &\highlight{2.18}  &9.748    \\
$\beta$-VAE~\cite{higgins2017beta}          & \highlight{0.0265} & 13.60    & \highlight{5.38}       & \highlight{1.22}  & \highlight{10.72}  & 11.33    &3.98   &7.705      \\
VampPrior VAE~\cite{tomczak2017vae}         & \highlight{0.0107} & 30.12    &\highlight{4.80}       & 2.10  & 14.44  & \highlight{5.97}    &4.20  & 10.270      \\
\textit{Mixup}-VAE                          & 0.0310 & 18.70    & 6.34       & 1.98  & 13.96  & 14.17    &3.06  &9.702   \\
\textit{Jigsaw}-VAE (ours)\ \ \ \                   & 0.0273 & \highlight{11.26}    & 6.00       & 1.60  & 14.10  & \highlight{10.27}    &\highlight{0.74}  &\highlight{7.328}        \\
\textit{Jigsaw}-$\beta$-VAE (ours)          & 0.0287 & \highlight{13.00}    & \highlight{5.38}       & \highlight{0.86}  & \highlight{9.96}  & 11.81    &2.46  &\highlight{7.252}          \\ \hline
\end{tabular}
\label{tab:celeb}
\end{table}

\begin{figure}[htb!]
    \centering
    \includegraphics[width=\textwidth]{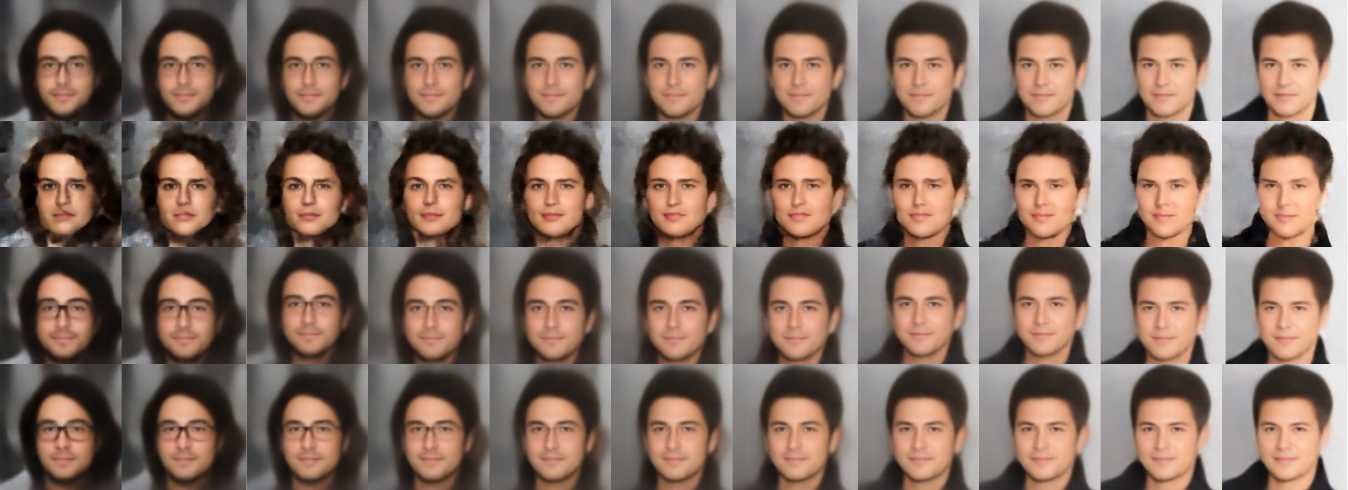}
    \caption{Transition from with to without eyeglasses. Rows from top to bottom correspond to VAE, VampPrior VAE, $\beta$-VAE, and \textit{Jigsaw}-VAE, respectively.}
    \label{fig:interpolation}
\end{figure}

We next examine the proposed method's ability to produce smooth interpolation (i.e. gradual transition). In Figure~\ref{fig:interpolation}, we show an interpolation sample between with eyeglasses and without eyeglasses. As shown in the third row of Figure~\ref{fig:interpolation}, \textit{Jigsaw}-VAE gradually removes the eyeglasses from the image, moving from left to right. However, in the VAE interpolation outputs as shown in the first row, the eyeglasses feature is suddenly disappeared after the second sample from left. The VampPrior VAE (second row), performs the worse by completely ignoring the eyeglasses attribute. $\beta$-VAE (the second row) encourages gradual transition compared to the VAE, however not as much as \textit{Jigsaw}-VAE does. 

\subsection{Quality of the Learned Latent Vectors in Clustering Context.}\label{subsec:clustering}

Compared to supervised models, their unsupervised counterparts are more susceptible to strong feature/distribution biases in the input data. That is, a dominant feature such as color or texture might confuse an unsupervised model to consider two different objects to be the same which have the same color/texture. In this section, in order to study the quality of the latent vectors in the VAE, we leverage the truncated Gaussian-mixture VAE~\cite{zhao2019variational}. We measure the clustering performance under an intended biased experiment; using colored MNIST dataset, we use a normal train set, but test with a biased set. 

\textbf{Colored MNIST.} For this experiment, we colorize each digit class in the MNIST~\cite{lecunmnisthandwrittendigit2010} train set with a specific color (Figure~\ref{fig:introfig}). To evaluate whether each method is able to capture more features besides the dominant \textit{color} feature, we use a single color code for all the test images. 

For this experiment, we compare the proposed \textit{Jigsaw}-VAE to four other methods. Similar to the generative task experiments in subsection~\ref{subsec:diversity}, we also train the VAE with the \textit{Mixup} approach. In Table~\ref{tab:mnist}, we report the results of clustering experiments. These results indicate that although applying $\beta$-VAE and \textit{d}-VAE improve generative task (subsection~\ref{subsec:diversity}), they do not improve the performance on the normal colored test set. However, the \textit{Mixup}-VAE and the proposed \textit{Jigsaw}-VAE improve the normalized mutual information (NMI)~\cite{song2019latent} score from $0.8872$ to $0.9022$ and $0.9473$, respectively. However, when we test the methods with intended color bias (Single-color results in Table~\ref{tab:mnist}), all the methods except for the \textit{Jigsaw}-VAE fail. This suggests that \textit{Jigsaw}-VAE is able to balance shape and color features, thus not completely failing when the test set is highly biased towards the color feature. For this experiment, the stochastic \textit{Jigsaw} permutation function runs over both spatial windows and RGB channels.

\begin{table}[htb!]
\centering
\caption{Colored MNIST clustering results. Multi-color refers to test set which follows the roles of the train set i.e. each digit is assigned the same specific color as the train set. Single-color refers to the experiment where we assign a single color to all digits in the test set. Higher values better.}
\begin{tabular}{lcc}
\hline
           & \multicolumn{2}{c}{NMI}    \\ \cline{2-3} 
Methods                                     & Multi-color & Single-color (median) \\ \hline
VAE~\cite{kingma2013auto}               & 0.8872      & 0.176      \\
\textit{d}-VAE~\cite{im2017denoising}       & 0.8713      & 0.041      \\
$\beta$-VAE~\cite{higgins2017beta}          & 0.8766      & 7.8E-06    \\
\textit{Mixup}-VAE                          & 0.9022      & 0.114      \\
\textit{Jigsaw}-VAE (ours)                        & \highlight{0.9473}      & \highlight{0.379}      \\ \hline
\end{tabular}
\label{tab:mnist}
\end{table} 

\begin{figure}[htb!]
     \centering
     \includegraphics[width=\textwidth]{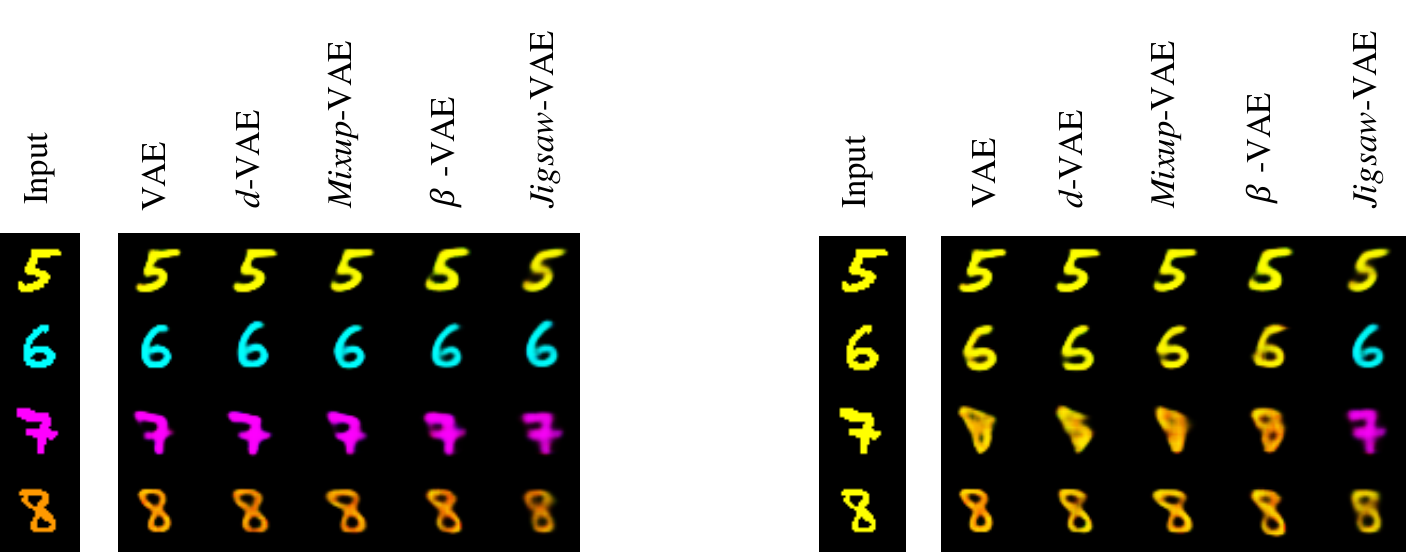}
     \caption{Colored MNIST reconstructed samples from decoding part of the clustering model~\cite{zhao2019variational}. Left: The input images (first column) have the same color codes as the train set distribution. Right: Input color codes are all mapped to a single code (here to yellow which is originally the color code for digit 5 in the train set) from train distribution.}
     \label{fig:mnistbias}
 \end{figure}

For this experiment, we adopt the truncated Gaussian-mixture VAE clustering approach~\cite{zhao2019variational} and modify it according to the different approaches used in the previous experiments i.e. VAE, \textit{d}-VAE, \textit{Mixup}-VAE, $\beta$-VAE, and \textit{Jigsaw}-VAE. Ideally, we want the clusters created by each method to be based on both color and structural information. If a model assigns a correct cluster label to an input, the decoded information of the input should contain both correct color and shape features. In Figure~\ref{fig:mnistbias}, we visualize the reconstructed digits using clustering models trained with colored MNIST. When the test set color codes are the same as the training digit colors, cluster assignments are fairly accurate (Table~\ref{tab:mnist}) thus reconstructions are reasonable for all the methods (Figure~\ref{fig:mnistbias} left). However, when all the digits in the test set take only one color code from train set, wrong clusters are assigned which results in wrong decodings (Figure~\ref{fig:mnistbias} right). As can be seen, only \textit{Jigsaw}-VAE is able to \textit{recover} the actual color codes that the model was trained with. Since color ``yellow'' is the actual color for digit ``$5$'', non-Jigsaw methods tend to wrongly reconstruct $5$-like images (second row, right) even though the input is digit ``$6$''. In another case (third row, right), non-Jigsaw approaches reconstruct a number form a closest color code to yellow, which in our case is ``$8$'' with color ``orange''. The results indicate that in the case of non-jigsaw VAE, color is learned as the dominant feature while the structural features are undermined. Where as the \textit{Jigsaw}-vae, balances the structural and color features.

\section{Conclusion}
We introduced \textit{Jigsaw}-VAE, a new VAE  capable of learning balanced features that can be used in both generative and other downstream tasks such as clustering. Our method exploits a mixture prior and a stochastic permutation layer. The mixture prior helps smoothing the conflict between the likelihood and KL terms while the permutation layer enforces the VAE to learn spatial relationships between the tiles (object parts). We empirically showed this prevents the VAE from ignoring infrequent spatial contexts (i.e. features). We showed that input permutation reduces the effect of strong damaging feature-bias in clustering models where we use VAEs as backbone. The \textit{Jigsaw}-VAE also showed smooth transitions during interpolation which is useful for preventing abrupt changes while modifying factors of a sample based on another sample.

Besides measuring reconstruction error, sampling quality and diversity, and how realistic the generated images are, we proposed to measure the presence of train data features in the generated images using our proposed metric, FPM. Among competing methods, $\beta$-VAE showed competitive performance on FPM and MSE metrics for the generative task (celebA experiments), however, it generates less realistic images compared to our \textit{Jigsaw}-VAE. Additionally, $\beta$-VAE showed sub-optimal performance for the clustering task (MNIST experiments) when test set is biased. Therefore, considering both generative and clustering tasks, our \textit{Jigsaw}-VAE outperformed competing methods.  

\bibliographystyle{plain}
\bibliography{neurips_2020}

\begin{thebibliography}{10}

\bibitem{asperti2020balancing}
Andrea Asperti and Matteo Trentin.
\newblock Balancing reconstruction error and kullback-leibler divergence in
  variational autoencoders.
\newblock {\em arXiv preprint arXiv:2002.07514}, 2020.

\bibitem{barratt2018note}
Shane Barratt and Rishi Sharma.
\newblock A note on the inception score.
\newblock {\em arXiv preprint arXiv:1801.01973}, 2018.

\bibitem{bowman2015generating}
Samuel~R Bowman, Luke Vilnis, Oriol Vinyals, Andrew~M Dai, Rafal Jozefowicz,
  and Samy Bengio.
\newblock Generating sentences from a continuous space.
\newblock {\em arXiv preprint arXiv:1511.06349}, 2015.

\bibitem{breiman2001random}
Leo Breiman.
\newblock Random forests.
\newblock {\em Machine learning}, 45(1):5--32, 2001.

\bibitem{brock2016neural}
Andrew Brock, Theodore Lim, James~M Ritchie, and Nick Weston.
\newblock Neural photo editing with introspective adversarial networks.
\newblock {\em arXiv preprint arXiv:1609.07093}, 2016.

\bibitem{chen2016variational}
Xi~Chen, Diederik~P Kingma, Tim Salimans, Yan Duan, Prafulla Dhariwal, John
  Schulman, Ilya Sutskever, and Pieter Abbeel.
\newblock Variational lossy autoencoder.
\newblock {\em International Conference on Learning Representations}, 2017.

\bibitem{clark2019don}
Christopher Clark, Mark Yatskar, and Luke Zettlemoyer.
\newblock Don't take the easy way out: Ensemble based methods for avoiding
  known dataset biases.
\newblock {\em arXiv preprint arXiv:1909.03683}, 2019.

\bibitem{dai2019diagnosing}
Bin Dai and David Wipf.
\newblock Diagnosing and enhancing vae models.
\newblock {\em arXiv preprint arXiv:1903.05789}, 2019.

\bibitem{deng2009imagenet}
Jia Deng, Wei Dong, Richard Socher, Li-Jia Li, Kai Li, and Li~Fei-Fei.
\newblock Imagenet: A large-scale hierarchical image database.
\newblock In {\em 2009 IEEE conference on computer vision and pattern
  recognition}, pages 248--255. Ieee, 2009.

\bibitem{fisher2019all}
Aaron Fisher, Cynthia Rudin, and Francesca Dominici.
\newblock All models are wrong, but many are useful: Learning a variable’s
  importance by studying an entire class of prediction models simultaneously.
\newblock {\em Journal of Machine Learning Research}, 20(177):1--81, 2019.

\bibitem{hardt2016equality}
Moritz Hardt, Eric Price, and Nati Srebro.
\newblock Equality of opportunity in supervised learning.
\newblock In {\em Advances in neural information processing systems}, pages
  3315--3323, 2016.

\bibitem{hendricks2018women}
Lisa~Anne Hendricks, Kaylee Burns, Kate Saenko, Trevor Darrell, and Anna
  Rohrbach.
\newblock Women also snowboard: Overcoming bias in captioning models.
\newblock In {\em European Conference on Computer Vision}, pages 793--811.
  Springer, 2018.

\bibitem{heusel2017gans}
Martin Heusel, Hubert Ramsauer, Thomas Unterthiner, Bernhard Nessler, and Sepp
  Hochreiter.
\newblock Gans trained by a two time-scale update rule converge to a local nash
  equilibrium.
\newblock In {\em Advances in neural information processing systems}, pages
  6626--6637, 2017.

\bibitem{higgins2017beta}
Irina Higgins, Loic Matthey, Arka Pal, Christopher Burgess, Xavier Glorot,
  Matthew Botvinick, Shakir Mohamed, and Alexander Lerchner.
\newblock beta-vae: Learning basic visual concepts with a constrained
  variational framework.
\newblock {\em Iclr}, 2(5):6, 2017.

\bibitem{im2017denoising}
Daniel Im~Jiwoong Im, Sungjin Ahn, Roland Memisevic, and Yoshua Bengio.
\newblock Denoising criterion for variational auto-encoding framework.
\newblock In {\em Thirty-First AAAI Conference on Artificial Intelligence},
  2017.

\bibitem{kim2019learning}
Byungju Kim, Hyunwoo Kim, Kyungsu Kim, Sungjin Kim, and Junmo Kim.
\newblock Learning not to learn: Training deep neural networks with biased
  data.
\newblock In {\em Proceedings of the IEEE Conference on Computer Vision and
  Pattern Recognition}, pages 9012--9020, 2019.

\bibitem{kim2018disentangling}
Hyunjik Kim and Andriy Mnih.
\newblock Disentangling by factorising.
\newblock {\em arXiv preprint arXiv:1802.05983}, 2018.

\bibitem{kingma2013auto}
Diederik~P Kingma and Max Welling.
\newblock Auto-encoding variational bayes.
\newblock {\em arXiv preprint arXiv:1312.6114}, 2013.

\bibitem{kingma2016improved}
Durk~P Kingma, Tim Salimans, Rafal Jozefowicz, Xi~Chen, Ilya Sutskever, and Max
  Welling.
\newblock Improved variational inference with inverse autoregressive flow.
\newblock In {\em Advances in neural information processing systems}, pages
  4743--4751, 2016.

\bibitem{lecunmnisthandwrittendigit2010}
Yann LeCun and Corinna Cortes.
\newblock {MNIST} handwritten digit database.
\newblock {\em http://yann.lecun.com/exdb/mnist/}, 2010.

\bibitem{leino2018feature}
Klas Leino, Emily Black, Matt Fredrikson, Shayak Sen, and Anupam Datta.
\newblock Feature-wise bias amplification.
\newblock {\em arXiv preprint arXiv:1812.08999}, 2018.

\bibitem{li2019repair}
Yi~Li and Nuno Vasconcelos.
\newblock Repair: Removing representation bias by dataset resampling.
\newblock In {\em Proceedings of the IEEE Conference on Computer Vision and
  Pattern Recognition}, pages 9572--9581, 2019.

\bibitem{li2018resound}
Yingwei Li, Yi~Li, and Nuno Vasconcelos.
\newblock Resound: Towards action recognition without representation bias.
\newblock In {\em Proceedings of the European Conference on Computer Vision
  (ECCV)}, pages 513--528, 2018.

\bibitem{liu2015deep}
Ziwei Liu, Ping Luo, Xiaogang Wang, and Xiaoou Tang.
\newblock Deep learning face attributes in the wild.
\newblock In {\em Proceedings of the IEEE international conference on computer
  vision}, pages 3730--3738, 2015.

\bibitem{lorenz2019unsupervised}
Dominik Lorenz, Leonard Bereska, Timo Milbich, and Bjorn Ommer.
\newblock Unsupervised part-based disentangling of object shape and appearance.
\newblock In {\em Proceedings of the IEEE Conference on Computer Vision and
  Pattern Recognition}, pages 10955--10964, 2019.

\bibitem{louizos2015variational}
Christos Louizos, Kevin Swersky, Yujia Li, Max Welling, and Richard Zemel.
\newblock The variational fair autoencoder.
\newblock {\em arXiv preprint arXiv:1511.00830}, 2015.

\bibitem{mathieu2018disentangling}
Emile Mathieu, Tom Rainforth, N~Siddharth, and Yee~Whye Teh.
\newblock Disentangling disentanglement in variational autoencoders.
\newblock {\em arXiv preprint arXiv:1812.02833}, 2018.

\bibitem{noroozi2016unsupervised}
Mehdi Noroozi and Paolo Favaro.
\newblock Unsupervised learning of visual representations by solving jigsaw
  puzzles.
\newblock In {\em European Conference on Computer Vision}, pages 69--84.
  Springer, 2016.

\bibitem{paumard2018jigsaw}
Marie-Morgane Paumard, David Picard, and Hedi Tabia.
\newblock Jigsaw puzzle solving using local feature co-occurrences in deep
  neural networks.
\newblock In {\em 2018 25th IEEE International Conference on Image Processing
  (ICIP)}, pages 1018--1022. IEEE, 2018.

\bibitem{razavi2019generating}
Ali Razavi, Aaron van~den Oord, and Oriol Vinyals.
\newblock Generating diverse high-fidelity images with vq-vae-2.
\newblock In {\em Advances in Neural Information Processing Systems}, pages
  14837--14847, 2019.

\bibitem{salimans2016improved}
Tim Salimans, Ian Goodfellow, Wojciech Zaremba, Vicki Cheung, Alec Radford, and
  Xi~Chen.
\newblock Improved techniques for training gans.
\newblock In {\em Advances in neural information processing systems}, pages
  2234--2242, 2016.

\bibitem{sandler2018mobilenetv2}
Mark Sandler, Andrew Howard, Menglong Zhu, Andrey Zhmoginov, and Liang-Chieh
  Chen.
\newblock Mobilenetv2: Inverted residuals and linear bottlenecks.
\newblock In {\em Proceedings of the IEEE conference on computer vision and
  pattern recognition}, pages 4510--4520, 2018.

\bibitem{santa2017deeppermnet}
Rodrigo Santa~Cruz, Basura Fernando, Anoop Cherian, and Stephen Gould.
\newblock Deeppermnet: Visual permutation learning.
\newblock In {\em Proceedings of the IEEE Conference on Computer Vision and
  Pattern Recognition}, pages 3949--3957, 2017.

\bibitem{sonderby2016ladder}
Casper~Kaae S{\o}nderby, Tapani Raiko, Lars Maal{\o}e, S{\o}ren~Kaae
  S{\o}nderby, and Ole Winther.
\newblock Ladder variational autoencoders.
\newblock In {\em Advances in neural information processing systems}, pages
  3738--3746, 2016.

\bibitem{song2019latent}
Tianbao Song, Jingbo Sun, Bo~Chen, Weiming Peng, and Jihua Song.
\newblock Latent space expanded variational autoencoder for sentence
  generation.
\newblock {\em IEEE Access}, 7:144618--144627, 2019.

\bibitem{suresh2019framework}
Harini Suresh and John~V Guttag.
\newblock A framework for understanding unintended consequences of machine
  learning.
\newblock {\em arXiv preprint arXiv:1901.10002}, 2019.

\bibitem{tolstikhin2017wasserstein}
Ilya Tolstikhin, Olivier Bousquet, Sylvain Gelly, and Bernhard Schoelkopf.
\newblock Wasserstein auto-encoders.
\newblock {\em International Conference on Learning Representations}, 2018.

\bibitem{tomczak2017vae}
Jakub~M Tomczak and Max Welling.
\newblock Vae with a vampprior.
\newblock {\em arXiv preprint arXiv:1705.07120}, 2017.

\bibitem{yeung2017tackling}
Serena Yeung, Anitha Kannan, Yann Dauphin, and Li~Fei-Fei.
\newblock Tackling over-pruning in variational autoencoders.
\newblock {\em arXiv preprint arXiv:1706.03643}, 2017.

\bibitem{zhang2017mixup}
Hongyi Zhang, Moustapha Cisse, Yann~N Dauphin, and David Lopez-Paz.
\newblock mixup: Beyond empirical risk minimization.
\newblock {\em arXiv preprint arXiv:1710.09412}, 2017.

\bibitem{zhao2017men}
Jieyu Zhao, Tianlu Wang, Mark Yatskar, Vicente Ordonez, and Kai-Wei Chang.
\newblock Men also like shopping: Reducing gender bias amplification using
  corpus-level constraints.
\newblock {\em arXiv preprint arXiv:1707.09457}, 2017.

\bibitem{zhao2019variational}
Qingyu Zhao, Nicolas Honnorat, Ehsan Adeli, Adolf Pfefferbaum, Edith~V
  Sullivan, and Kilian~M Pohl.
\newblock Variational autoencoder with truncated mixture of gaussians for
  functional connectivity analysis.
\newblock In {\em International Conference on Information Processing in Medical
  Imaging}, pages 867--879. Springer, 2019.

\end{thebibliography}

\end{document}